\documentclass[11pt]{article}

\usepackage[utf8]{inputenc}
\usepackage[T1]{fontenc}
\IfFileExists{lmodern.sty}{\usepackage{lmodern}}{}
\usepackage[margin=1in]{geometry}
\usepackage{amsmath,amssymb}
\usepackage{graphicx}
\usepackage{float}
\usepackage{booktabs}
\usepackage{array}
\usepackage{caption}
\usepackage{subcaption}
\usepackage{authblk}
\usepackage{textcomp}
\IfFileExists{microtype.sty}{\usepackage{microtype}}{}
\usepackage{enumitem}
\usepackage[hidelinks,breaklinks=true]{hyperref}
\usepackage{xcolor}
\usepackage{tabularx}
\newcolumntype{R}{>{\raggedleft\arraybackslash}X}

\title{\textbf{Loggia dei Lanzi: AI Thermography Enhancement\\
Comparisons through 3D Photogrammetry}}

\author[1]{Scott McAvoy}
\author[1]{Jonathan Klingspon}
\author[2]{George Bent}
\author[2]{Dave Pfaff}
\author[1]{Aviral Agarwal}
\author[3]{Maurizio Seracini}
\author[1]{Falko Kuester}

\affil[1]{Cultural Heritage Engineering Initiative (CHEI), University of California San Diego}
\affil[2]{Washington and Lee University}
\affil[3]{Editech, Florence, Italy}

\date{}

\begin{document}
\maketitle

\begin{center}
\vspace{-1.5em}
\textit{Preprint. Submitted to the 2026 International Symposium on Cultural Heritage
Conservation by Digitization (CHCD 2026).}
\end{center}

\begin{abstract}
\noindent
The Loggia dei Lanzi in the Piazza della Signoria is one of Florence's most prominent
structures visited by millions every year. Its construction history spans multiple centuries
of modification. This paper presents the results of a thermal imaging campaign conducted in
December 2025, using a FLIR T1020 HD camera, revealing hidden architectural
features including walled-up openings and material transitions beneath the plaster surface.
The favorable winter ambient conditions provided a feature-rich benchmark upon which to compare the
results of enhancement algorithms and artificial intelligence models. We evaluate the
application of AI-based image enhancement to thermal heritage documentation through a
comparison of three tiers of image resolution in a photogrammetric Structure-from-Motion (SfM)
pipeline: native resolution, FLIR's hardware-based pixel-shifted super-resolution (UltraMax),
and state of the art AI-upscaled imagery models. We quantify the effect of each resolution tier
on feature detection and tie-point generation, assessing whether the additional detail produced
by super-resolution, whether hardware or AI-derived, translates into meaningfully denser and
more accurate 3D thermal models. Our results contribute to the emerging intersection of
artificial intelligence and heritage thermography by providing a direct comparison of hardware
microscanning and AI super-resolution within a thermal photogrammetric workflow for cultural
heritage. All datasets are made publicly available and accessible within an interactive 3D
archival framework, and integrated into a custom citywide extended reality overlay application.

\vspace{0.5em}
\noindent\textbf{Keywords:} Thermography, AI enhancement, Super-resolution, Florence, XR
\end{abstract}

\section{The Loggia dei Lanzi}

Whereas written records of fourteenth-century artistic projects are often scarce, Florentine
institutions were quite careful when it came to documenting the progress of architectural
projects in the city. The selection of masters, financial supervisors, and specialists in
charge of a building's specific features were of great importance to patrons and institutional
leaders charged with fiduciary responsibilities. The aggregation of payments to named
individuals, along with reprimands and pecuniary threats to laborers perceived to be less than
faithful to their duties, can paint a vivid picture of a structure's building history. As Carl
Frey demonstrated masterfully in 1885 [1], such is the case with the evidence surrounding the
construction of the Loggia dei Lanzi, the large public meeting area that faces the Palazzo
Vecchio from the south flank of the Piazza della Signoria, located in the civic heart of
Florence.

\begin{figure}[H]
\centering
\includegraphics[width=\linewidth]{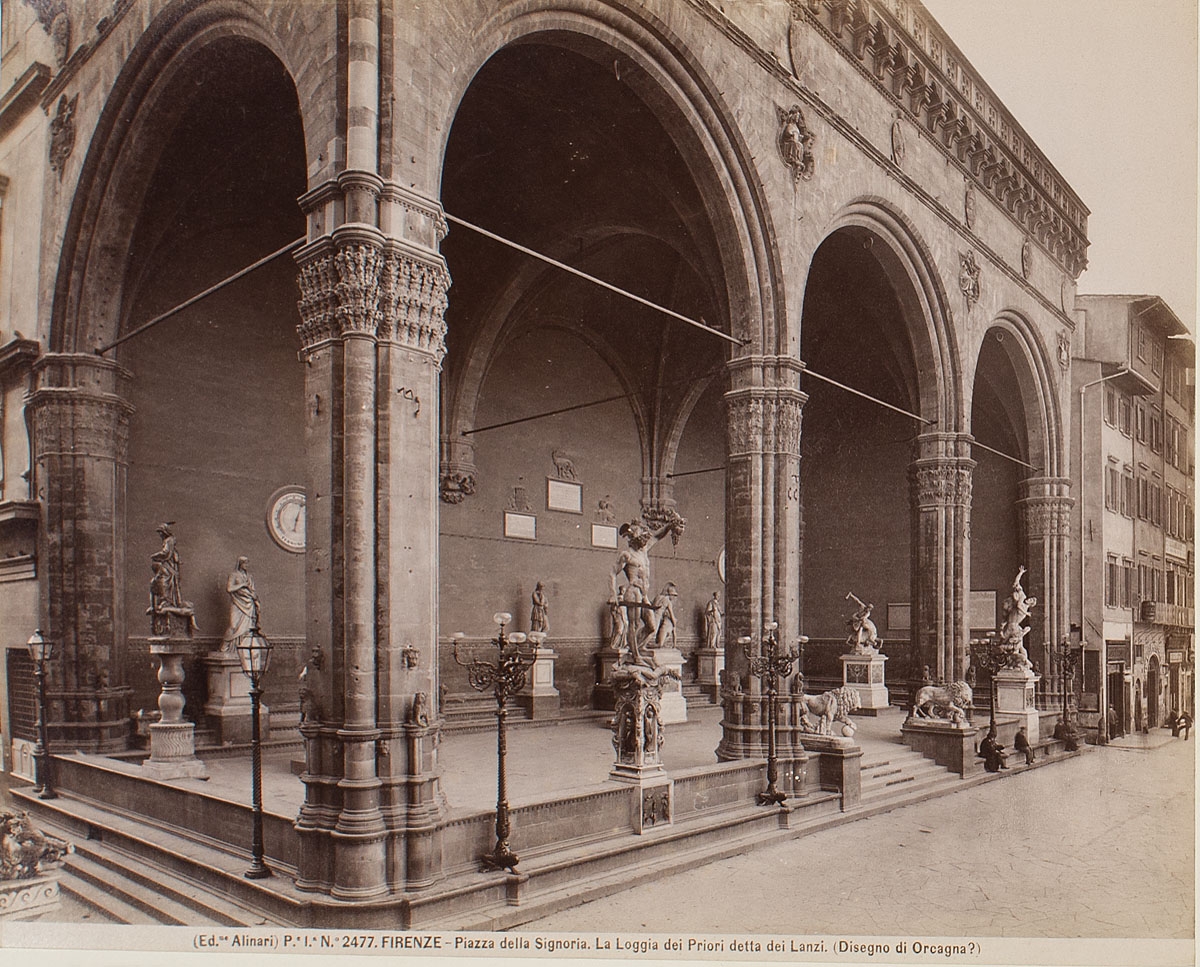}
\caption{Image of Loggia in late 19th century (Source: Fratelli Alinari archive).}
\label{fig:loggia_alinari}
\end{figure}

The loggia form was a common feature of public life in the Medieval city. Used primarily as an
architectural framing device for performative celebrations but also employed as a feature of
markets and food centers, the open design relied on the simple combination of piers, bases, and
vaulted ceilings while omitting solid wall surfaces. As a built covering, the loggia served as a
protected space in which users could shield themselves from the elements --- rain, snow, sleet,
and sunshine --- all year round. The permeable structure encouraged free movement into and
through the space, and benches were often placed inside and outside the canopied loggia to
foster conversations both formal and informal. Loggias could come in any size or shape, and it
was not unusual for particularly affluent Florentine families to construct such an architectural
space either in or opposite their urban residences as a kind of public waiting room, so that
clients --- both commercial and social --- could congregate there, for all to see, in the hopes
of meeting with their patron (Figure \ref{fig:loggia_alinari}).

Florentine officials recognized the importance of these functional aspects when the city's
priors formally moved to commission the construction of the large loggia on January 14, 1374.
Citing the need for a space large enough to hold official ceremonies, the move to purchase
properties on the Piazza della Signoria directly across from the Palazzo Vecchio --- the
physical center of the Republican government --- was overtly motivated by a desire to enhance the
prestige of the commune as an enormous stage upon which to perform acts of authority, both local
and international. Buildings along the southern edge of the piazza were purchased from their
owners and then demolished, creating an open space that abutted the city's Mint, located just to
the south of the building site, thus connecting an important administrative office to the Palazzo
Vecchio via the intended loggia.

Perhaps predictably, the desires of the priors outpaced their means as the proclamation to build
the loggia was not acted upon for three full years. Only in 1377, after securing the
participation (and oversight) of the supervisors then employed by the nearly completed cathedral
of S. Maria del Fiore, was a construction team appointed, salaried, and charged with specific
building and decorative tasks. Not surprisingly, the final designs of the piers and vaults of the
Loggia dei Lanzi referred directly to those that had just been erected in the Duomo during its
construction in the middle decades of the fourteenth century. Among those involved in the
architectural and artistic project were Simone Talenti and Benci di Cione, experienced and
talented specialists in the arts of sculpture and building. Neither of them appears to have been
in any hurry to complete the work, as the rather basic triple arched loggia (and the sculptural
reliefs embedded in it) took five full years to complete. It was inaugurated as a usable space by
the priors on 01 November 1382 and then put into service only four days later when diplomats from
Ravenna were officially received there.

The simple plan of the Loggia dei Lanzi reveals the limitations placed upon its designers from
the project's outset. The solid walls along its southern and western flanks abutted extant
structures --- most importantly the offices of the City Mint --- that required architectonic
protection from people and the elements. Three large arches formed by two wide and tall piers
articulate the loggia's east and north sides that face the piazza and, by implication, the
Palazzo Vecchio. Vast combinations of acanthus leaves, faint vestiges of classical architectural
capital designs, ornament these octagonal piers. From these piers spring ribbed vaults that reach
up into the ceiling before descending to a pair of corbels attached to the south wall. The design
borrows heavily from Florentine prototypes in the Duomo, the churches of S. Croce and S. Maria
Novella, and the nearby granary/cult center of Orsanmichele.

Multiple lions' heads, early representations of the symbol of Florence (and, perhaps, allusions
to the live lions kept at the Palazzo Vecchio by the government), adorn the lower sections of
these octagonal piers, while crouching human figures --- perhaps Adam and Eve flanked by either
angels or their two sons, Cain and Abel --- support on their hunched shoulders the corbels from
which spring the ribs of the south vault. The exterior spandrels, visible from the piazza, contain
sculpted reliefs of the four Cardinal Virtues --- Fortitude, Justice, Temperance, and Prudence ---
carved between 1383 and 1387 by Jacopo di Piero Guidi and Giovanni d'Ambrogio according to Agnolo
Gaddi's designs, for which the latter artist was compensated in 1383 [1]. Below them appear
severely deteriorated sculpted shields bearing the coats of arms of Florence's allies and
protectors --- the House of Anjou, the Papacy, the Guelph Party, the Popolo of Florence, etc. The
entire complex was designed to showcase the authority of the Republican government of the city.

\section{Thermographic Imaging for Cultural Heritage}

Infrared thermography (IRT) is a well-established non-destructive testing (NDT) technique in
heritage science. Pulsed and passive thermography have seen increasing use for the depth-resolved
detection of subsurface features in documentary materials, panel paintings, mosaics, and masonry
[2]. Thermographic surveys of historic buildings exploit the principle that structural elements of
varying thickness or material composition produce differential thermal patterns when subject to a
continuous thermal gradient through the building envelope. Winter conditions are therefore
preferred for passive surveys, as the temperature differential between heated interiors and cold
exteriors maximizes contrast [3]. Lehmann et al.\ demonstrated that hidden structural elements ---
including walled-up windows and doors --- can be reliably located via steady-state infrared
thermography of historic masonry walls captured during winter months [4], an approach directly
analogous to the methodology employed in the present study. Recent reviews chronicle the
methodological maturation of infrared thermography for built heritage --- spanning advances in
acquisition geometry, survey design, and the integration of thermographic data with photogrammetric
and range-based 3D models --- consolidating IRT's transition from a qualitative diagnostic aid to a
metrically grounded documentation technique [36].

Deep learning has recently been introduced into thermographic monitoring of cultural heritage.
Garrido et al.\ presented the first integration of deep learning and IRT for heritage inspection,
developing automatic thermogram pre-processing algorithms that improved defect detection accuracy
[5]. Subsequent work has applied convolutional neural networks (CNNs) to the automated
classification of rising damp [6] and moisture detection [7] from thermograms of historic masonry.
Zhou et al.\ designed a digital conservation platform integrating thermal infrared imaging with
multimodal data fusion and AI-based classification algorithms, demonstrating improved recognition
of surface details and damage patterns on cultural heritage surfaces [8].

\begin{figure}[H]
\centering
\includegraphics[width=\linewidth]{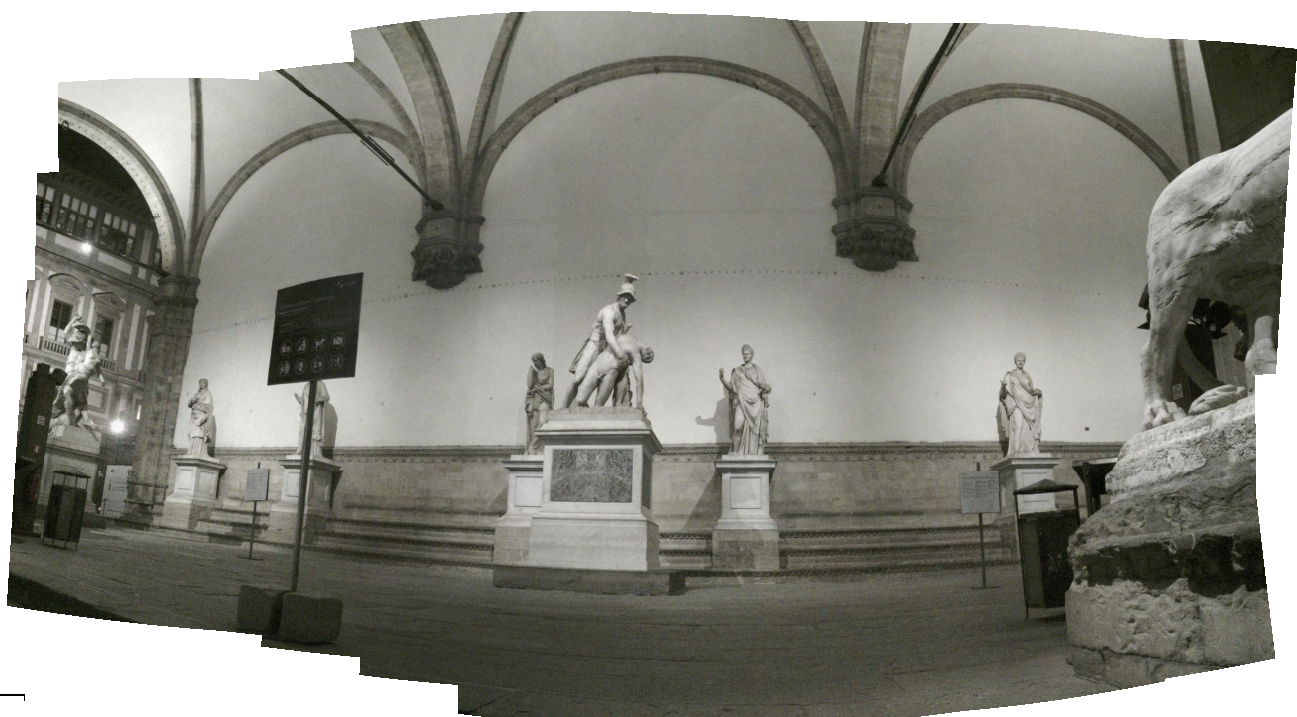}\\[6pt]
{\small (\textbf{a}) }\\[10pt]
\includegraphics[width=\linewidth]{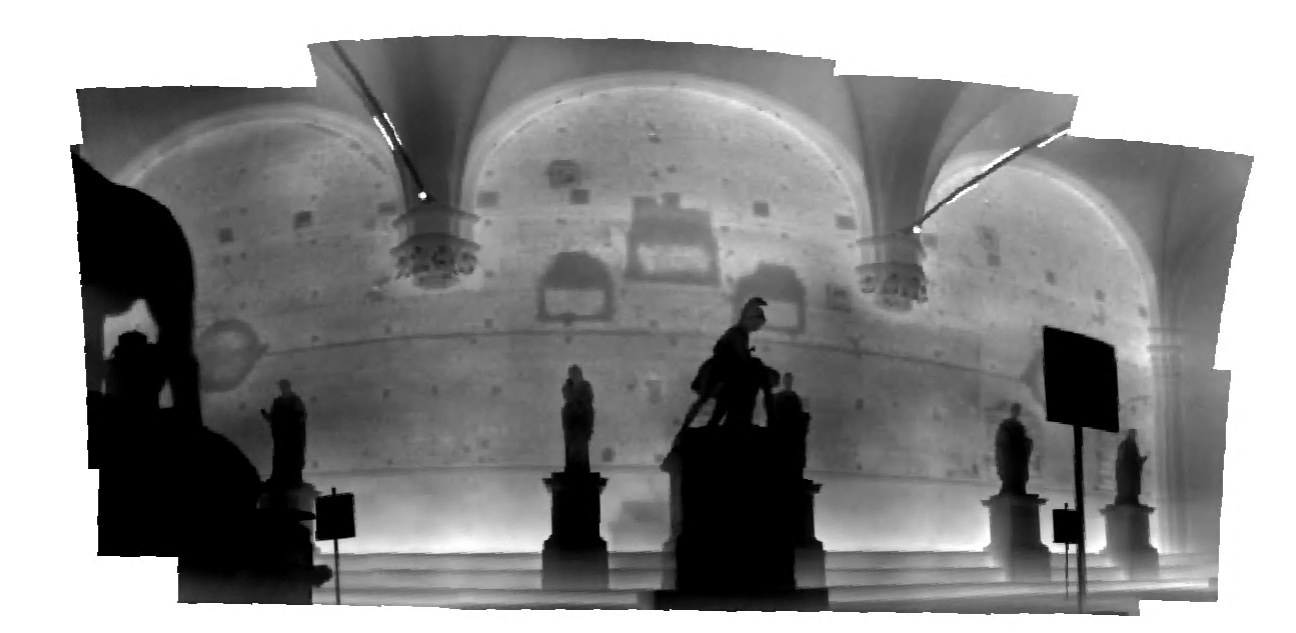}\\[6pt]
{\small (\textbf{b}) }
\caption{Visible light panoramic mosaic (a) and thermographic mosaic (b) showing hidden architectural features}
\label{fig:twopanel}
\end{figure}

In Florence specifically, thermography has a long and significant legacy. Maurizio Seracini
pioneered the use of multispectral imaging and diagnostic technologies applied to works of art and
historic buildings [9], including the use of thermography to discover hidden windows in the Palazzo
Vecchio's Salone dei Cinquecento. The development of augmented reality systems for overlaying
diagnostic imaging data onto heritage structures --- from the original ARtifact tablet-based system
[10] through the current citywide implementation tested at the Loggia dei Lanzi and other Florentine
monuments [11] --- provides a direct pathway from thermographic capture to public engagement with
hidden heritage (Figure \ref{fig:twopanel}).

\section{Thermal Mosaic and 3D Thermographic Reconstruction}

The creation of large-scale thermal mosaics from individual camera frames involves challenges
distinct from visible-light panorama stitching. Thermal images typically exhibit lower spatial
resolution, reduced contrast, and fewer distinctive features for image matching compared to RGB
equivalents. Previtali et al.\ developed a methodology for mapping IR thermal images onto 3D models
created with terrestrial laser scanning, employing rigorous photogrammetric orientation of both
thermal and RGB images in a combined bundle adjustment [12]. Parisotto et al.\ addressed the
complementary problem of light inhomogeneity in thermal mosaics using osmosis filtering [13]. The
FLIR T1020 HD's built-in panorama function produces a first-order mosaic, but achieving metrically
accurate, radiometrically consistent mosaics for analytical purposes requires post-processing
registration, often guided by co-acquired RGB imagery or LiDAR geometry.

\section{Super-Resolution for Thermal Images}

The resolution gap between thermal and visible-light sensors has driven substantial research into
super-resolution (SR) for infrared imagery, encompassing both hardware-based and computational
approaches.

Hardware-based SR through FLIR's UltraMax system is a commercial implementation of multi-frame
microscanning, a technique with well-established foundations in sampling theory [14, 15, 16].
Infrared focal plane arrays are typically undersampled relative to their optical resolution, so
spatial frequencies above the detector's Nyquist limit are aliased rather than resolved.
Microscanning addresses this by capturing several frames at slightly different sub-pixel phases;
because each frame samples the scene differently, the combined set contains higher
spatial-frequency information than any single frame, which can be recovered through registration
onto a finer grid. Critically, because the recovered detail is constrained by multiple genuine
observations rather than learned priors, multi-frame methods recover real measured information
rather than the ``hallucinated'' structures that can arise in single-image approaches [17].
UltraMax captures sixteen frames in under one second and reconstructs an image with twice the
linear resolution, quadrupling the pixel count from $1024 \times 768$ to $2048 \times 1536$ [18].
FLIR's published material claims that radiometric data is preserved intact and that measurement
accuracy improves through the reduced effective spot size [18]. However, UltraMax relies on the
uncontrolled micro-tremor of the operator's hand to generate its sub-pixel offsets, rather than
the precise piezoelectric actuation used in laboratory microscanning systems; the offsets are
estimated post hoc rather than known a priori. Despite the technology's ubiquity in professional
thermography, no independent peer-reviewed study has quantified how much genuine detail FLIR's
proprietary reconstruction recovers relative to native resolution, leaving the manufacturer's
claims theoretically plausible but methodologically unverified. The present study provides, to the
authors' knowledge, the first such independent evaluation.

Computational single-image SR has progressed rapidly in the deep learning era, challenged by the
distinct properties of thermal images: low contrast, limited high-frequency details, and
sensor-specific noise [19]. CNN-based models have achieved substantial gains over traditional
interpolation [20]. The Swin Transformer architecture [21], which introduced shifted-window
self-attention, was adapted for image restoration by Liang et al.\ as SwinIR [22], achieving strong
PSNR and SSIM performance. Hsu et al.\ extended this approach with the Dense-Residual-Connected
Transformer (DRCT), which mitigates information bottlenecks through dense-residual connections
between transformer layers [23]. The DRCT architecture, fine-tuned on thermal imagery, won first
place in Track 1 (single thermal image SR, $\times 8$) of the Sixth Thermal Image Super-Resolution
Challenge at the PBVS workshop at CVPR 2025 [24], achieving a PSNR of 28.52 and SSIM of 0.8466. Li
et al.\ introduced DifIISR, a diffusion model with gradient guidance designed specifically for
infrared image super-resolution, incorporating thermal spectral distribution regulation to preserve
thermal characteristics [25]. DifIISR represents the current state of the art in perceptual quality
for thermal SR, though its diffusion-based generative approach prioritizes visual fidelity over
strict radiometric accuracy.

\section{AI Upscaling for Photogrammetric Reconstruction}

The question of whether AI-upscaled images produce meaningfully improved photogrammetric
reconstructions, measured by tie-point density, point cloud completeness, and geometric accuracy,
is an emerging research area. Parrinello et al.\ investigated the contribution of AI-based image
upscaling in the Structure-from-Motion (SfM) data acquisition process [26]. Pashaei et al.\ applied
ESRGAN to super-resolve low-resolution UAS imagery by a factor of four, confirming that imaging
geometry could be reliably retrieved from super-resolved image sets [27]. However, virtually all
related work has focused exclusively on visible-light imagery. The application of AI upscaling to
thermal SfM pipelines for heritage documentation remains unstudied, a gap the present paper aims to
address.

\section{Key Gaps Addressed by the Present Study}

The literature reveals several gaps that the present study is positioned to fill. First, no
published study has directly compared hardware-based microscanning SR (FLIR UltraMax) against
AI-based SR for heritage thermography, nor independently verified the effectiveness of UltraMax
against a non-informative interpolation baseline. Second, while AI upscaling has been evaluated in
visible-light photogrammetry, its application to thermal SfM pipelines for heritage documentation
is unstudied. Finally, the
Loggia dei Lanzi's rear wall, with its multi-period construction history, has not been the subject
of published thermographic investigation.

\section{Materials and Methods}

\subsection{Field Campaign}

Thermal imaging of the Loggia dei Lanzi was conducted on the evening of Saturday, December 6, 2025
as part of a broader thermographic survey of Florence's historic center [28]. Ambient conditions
were recorded at approximately 11\textdegree C with 43\% humidity by the on-site thermometer, though
online weather stations reported 6\textdegree C and 98\% humidity, a discrepancy noted in the field
log. The survey focused on the Loggia's rear (south) wall and the interior vault surfaces, capturing
panoramic sequences from within the loggia structure (Figure \ref{fig:twopanel}). A total of 161 thermal images were acquired across four tripod stations.

\begin{figure}[!htbp]
\centering
\includegraphics[width=\linewidth]{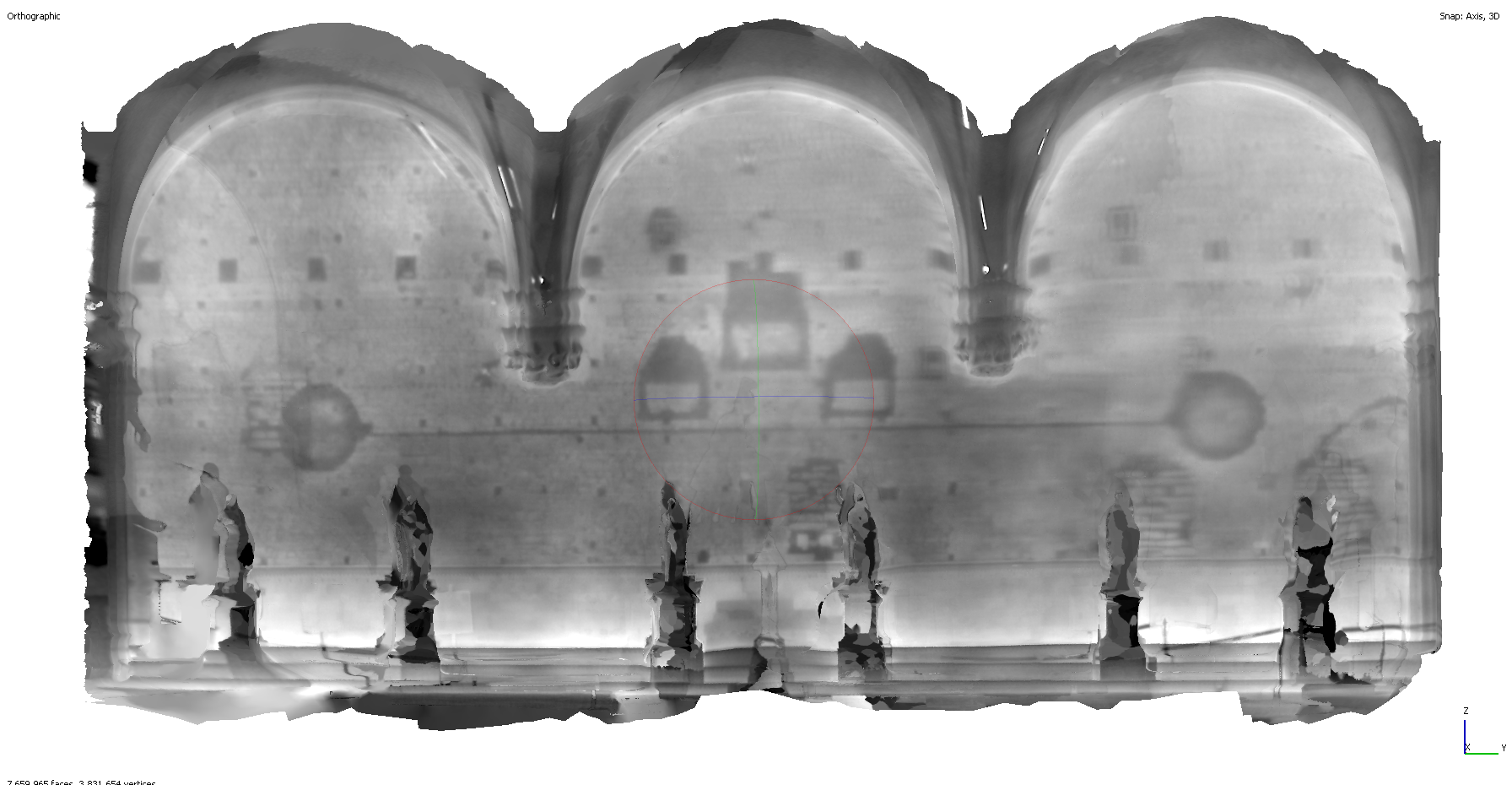}
\caption{3D orthomosaic of all native thermographic inputs, showing the Loggia dei Lanzi back wall.}
\label{fig:loggia_3d_result}
\end{figure}

All images were captured using a rented FLIR T1020 HD camera ($1024 \times 768$ native resolution,
$<$20~mK NETD at 30\textdegree C) with the UltraMax pixel-shifted super-resolution mode enabled,
producing co-registered native ($1024 \times 768$) and UltraMax ($2048 \times 1536$) outputs for
each capture (Figure \ref{fig:loggia_3d_result}). Because UltraMax exploits the small, involuntary micro-tremor of the handheld operator
to generate the sub-pixel frame-to-frame displacements its reconstruction requires [18], stationary
tripod mounting would in principle defeat the mechanism by eliminating that motion. To reconcile the
stability needed for station-based registration with the displacement UltraMax depends upon,
captures were made with a deliberately loosened tripod head, permitting slight on-center jiggle
while maintaining the nominal station position. This provided controlled sub-pixel displacement
analogous to hand tremor while preserving the geometric stability required for registration.

The thermal dataset was spatially registered to three Leica RTC 360 terrestrial LiDAR scans of the
Piazza della Signoria performed in April 2025 and published as an open dataset [29]. The LiDAR
survey provided an independent geometric reference frame to which all camera positions were aligned
and scaled, allowing reconstruction quality to be evaluated against a common, externally determined
geometry rather than a self-consistent but arbitrary photogrammetric coordinate system. Within this
reference frame, the mean horizontal standoff from the camera stations to the rear wall was
approximately 16.0~m, derived from the LiDAR-registered coordinates of the capture position and the
target wall surface; this standoff, together with the sensor resolution and lens geometry, fixes the
ground sample distance used to normalize the photogrammetric metrics (Section~\ref{sec:photo}).

\subsection{Image Extraction and Preparation}

Raw radiometric data was extracted from the FLIR proprietary format using the FLIRextractor tool
described in [31], producing 16-bit single-channel PNG files in which each pixel value encodes
calibrated temperature data. This 16-bit radiometric representation is critical for heritage
thermography, as it preserves the full dynamic range of temperature measurement across the scene,
where subsurface architectural features are revealed by differentials of a fraction of a degree.
All processing maintained this 16-bit depth through export, upscaling, and re-import into the
photogrammetric pipeline.

The decision to preserve 16-bit radiometric depth, rather than converting to 8-bit imagery as
required by most AI super-resolution models, was deliberate. The PBVS Thermal Image Super-Resolution
Challenge [24] and its competing architectures operate on 8-bit data derived from surveillance-class
thermal sensors (FLIR Tau2, $640 \times 480$), reflecting an orientation toward object detection and
human legibility rather than radiometric measurement. Our FLIR T1020 HD produces 16-bit radiometric
data at higher native resolution from a fundamentally different sensor class. To apply these models
without permanently discarding temperature information, each upscaling pipeline normalizes the 16-bit
thermal values to float32, processes them through the model, and rescales the output to the original
16-bit temperature range; roundtrip radiometric fidelity (mean absolute error, RMSE, and maximum
error against the native input) was recorded at each stage.

\subsection{Super-Resolution Methods}

Six resolution tiers were compared. The native tier and the manufacturer's hardware enhancement
establish, respectively, an unenhanced baseline and a non-AI reference; bicubic interpolation
provides a non-informative control; and three contemporary AI models span the range of current
approaches from domain-agnostic to domain-specific and from deterministic to generative.

\begin{enumerate}[leftmargin=*,itemsep=2pt]
  \item \textbf{Native} ($1024 \times 768$): raw extracted 16-bit thermal images at sensor
        resolution.
  \item \textbf{UltraMax} ($2048 \times 1536$): hardware multi-frame microscanning produced
        in-camera by the FLIR T1020 HD [18].
  \item \textbf{Bicubic} ($2048 \times 1536$): bicubic upscaling of the native images to UltraMax
        dimensions, distinguishing gains attributable to additional pixels from those attributable
        to genuine detail recovery.
  \item \textbf{SwinIR} ($2048 \times 1536$): the classical Swin Transformer SR model of Liang
        et al.\ [22], pretrained on the DIV2K natural-image dataset and applied without thermal
        fine-tuning, testing whether an off-the-shelf, PSNR-oriented model can enhance thermal
        photogrammetry.
  \item \textbf{DifIISR} ($4096 \times 3072$): the diffusion model of Li et al.\ [25], designed for
        infrared SR and presented at CVPR 2025, incorporating thermal spectral distribution
        regulation but optimizing for perceptual quality, with attendant risk of introducing
        fabricated detail.
  \item \textbf{TongJi-SR / DRCT} ($8192 \times 6144$): the first-place solution from the PBVS 2025
        Thermal Image Super-Resolution Challenge Track 1 [24], based on the DRCT architecture [23]
        fine-tuned on thermal imagery and operating at its native $\times 8$ scale.
\end{enumerate}

The tiers therefore span four distinct output resolutions --- $1\times$, $2\times$, $4\times$, and
$8\times$ the native dimension --- a fact that governs the normalization of all
resolution-dependent metrics (Section~\ref{sec:photo}). All AI models were executed on a single
NVIDIA GeForce RTX 4090 GPU (24~GB VRAM). Custom Python wrapper scripts managed the 16-bit thermal
roundtrip conversion for each model; these scripts and the full processing configuration are
provided in the Appendix.

\subsection{Photogrammetric Evaluation}
\label{sec:photo}

Each tier was processed through an identical Structure-from-Motion pipeline in Agisoft Metashape
Professional (v2.2). To isolate image content as the sole variable, two constraints were imposed
across all six projects. First, camera positions were imported as fixed reference coordinates from
the prior LiDAR-aligned project rather than re-estimated, so that network geometry was held
constant. Second, a fixed camera calibration was applied to each tier, derived by solving the
calibration once on the native imagery and scaling the focal length and principal-point offset by
the resolution ratio of each tier ($2\times$, $4\times$, or $8\times$); the dimensionless radial and
tangential distortion coefficients are resolution-independent and were transferred unchanged. Fixing
the calibration prevents per-tier self-calibration drift from confounding the comparison, so that
differences in the resulting sparse clouds reflect only the imagery.

Of the 161 images acquired, 160 entered the network; the number aligning successfully varied by tier
(from 154 to 159), itself a measure of how each enhancement method affects feature matchability and
reported as a result. To enable a strictly controlled comparison of feature richness and quality,
all tiers were then restricted to the intersection of images that aligned successfully in every tier
--- a common set of 149 images --- so that tie-point and projection counts are summed over an
identical network in every project.

For each tier, two classes of metric were recorded. Project-level statistics (tie-point count, total
projections, mean reprojection error, and ground sample distance) were taken from the Metashape
processing report. Per-point statistics were exported for every tie point via a custom Metashape
Python script (provided in the Appendix), yielding, for each point, its 3D coordinate, the number of
images in which it was observed, its reprojection error, its reconstruction uncertainty, and its
projection accuracy. These per-point exports allow the distribution of tie-point quality within each
cloud to be characterized rather than reduced to a single project mean.

\paragraph{Resolution normalization.}
Because the comparison spans four resolutions, pixel-based metrics are not directly comparable: a
pixel subtends a different physical distance at each tier. Reprojection error was therefore
normalized to physical units (millimeters on the wall surface) using the ground sample distance
(GSD). Rather than adopt Metashape's reported ground resolution --- which is back-computed from each
tier's reconstructed point distribution and therefore varies with cloud quality rather than
acquisition geometry --- GSD was derived analytically from the fixed acquisition geometry. With
camera positions, focal length, and target standoff held constant, the true GSD is a function of
resolution alone: 8.75~mm/px at native resolution, halving with each doubling of linear resolution to
4.375~mm/px ($2\times$ tiers), 2.188~mm/px (DifIISR), and 1.094~mm/px (TongJi). Reconstruction
uncertainty, by contrast, is a dimensionless ratio describing the elongation of each point's
triangulation error ellipsoid; because it depends on triangulation geometry rather than pixel size,
and because camera positions are identical across all tiers, it is directly comparable between tiers
without rescaling and serves as the principal discriminator of tie-point quality. Projection
accuracy, being reported in pixel units, was found to flatten to near-identical values across tiers
once divided by the resolution ratio and is consequently not used as a discriminating metric.

This framework tests not merely whether an upscaling method produces visually sharper images, but
whether the recovered detail is geometrically consistent across viewpoints --- a more rigorous
criterion than single-image quality metrics such as PSNR or SSIM. SIFT-based feature detection is
sensitive to local gradient structure; if a model introduces plausible but fictitious texture, that
detail will triangulate poorly and fail to localize consistently when matched across overlapping
images captured from different positions, manifesting as elevated reconstruction uncertainty. The
bicubic baseline controls for the possibility that increased tie-point counts arise simply from
additional pixels rather than from genuine detail recovery.

Each model was evaluated at its native output resolution rather than resampled to a common grid.
This choice was deliberate. A natural expectation is that all enhanced tiers should be resampled to
a single shared resolution before comparison --- most plausibly the $2\times$ ($2048 \times 1536$)
grid of the UltraMax hardware reference, so that every method is judged on an identical pixel
lattice. We deliberately did not do this. Resampling all outputs to a shared resolution would act as
a low-pass filter, suppressing precisely the high-frequency synthesized content whose geometric
consistency is under investigation; an aggressively upscaled image downsampled back to the
$2\times$ UltraMax grid would have its fabricated detail averaged away, masking rather than measuring
the very behavior of interest. Such a procedure would, in effect, pre-empt the experiment's central
question by erasing the synthesized structure before it could be tested for multi-view consistency.
Moreover, each model's native output resolution is its designed operating point --- the scale at
which it was trained and, in the case of the challenge-winning DRCT model, benchmarked --- so
evaluating it below that scale would assess a degraded variant rather than the method as deployed.
We instead retain each tier at its native scale and reconcile the differing resolutions through the
explicit physical-unit (GSD) normalization described above, supported by two resolution-independent
metrics (reconstruction uncertainty and multi-view consistency) that require no rescaling at all and
therefore cannot be biased by the resolution disparity. The consequence of this choice is a
comparison spanning four resolutions, which we address through normalization rather than by altering
the model outputs.

\subsection{Data Availability and Visualization}

All thermal datasets, including the 16-bit radiometric images and derived point clouds, are
published as open datasets on OpenHeritage3D [32] and are accessible through an interactive 3D web
viewer supporting direct query of per-point temperature values [33]. The Loggia dei Lanzi
thermography dataset [33] and the Palazzo Vecchio LiDAR reference scans [29] are available under
Creative Commons licenses. The thermal point-cloud models are additionally integrated into the
ARtifact extended reality application [10, 11], enabling in-situ overlay of thermographic data onto
the physical structure via mobile device. All processing scripts --- the six super-resolution
wrappers, the Metashape tie-point export script, and the analysis code generating the reported
statistics and figures --- together with the per-tier tie-point metric exports, are provided in the
Appendix and the associated public code repository [35].

\section{Results}

All six resolution tiers were processed through an identical Structure-from-Motion network of 149
images --- the subset that aligned successfully across every tier --- with camera positions locked
to the LiDAR-derived reference frame and a fixed, resolution-scaled calibration applied to each tier
(Section~\ref{sec:photo}). This ensures that differences in the resulting sparse clouds reflect image
content alone rather than variation in network geometry or self-calibration. Table~\ref{tab:main}
consolidates the principal metrics.

\begin{table}[ht]
\centering
\caption{Photogrammetric comparison across resolution tiers (common 149-image network). Reprojection
error is reported both in pixels (as output by Metashape) and normalized to millimeters on the wall
surface using the analytically derived ground sample distance (GSD). Reconstruction uncertainty (RU)
is a dimensionless triangulation-geometry ratio, reported here as the fraction of tie points
exceeding a threshold of 100.}
\label{tab:main}
\small
\begin{tabularx}{\textwidth}{l c R R R R R R}
\toprule
Tier & Resolution & Tie pts & Obs./pt & Reproj.\ (px) & Reproj.\ (mm) & RU $>$ 100 & $\geq$3 img \\
\midrule
Native    & $1024\times768$  & 22{,}550 & 3.48 & 1.85  & 16.2 & 0.0\%  & 45.5\% \\
UltraMax  & $2048\times1536$ & 20{,}660 & 3.41 & 4.28  & 18.7 & 7.8\%  & 42.7\% \\
Bicubic   & $2048\times1536$ & 19{,}484 & 2.87 & 3.88  & 17.0 & 41.0\% & 32.0\% \\
SwinIR    & $2048\times1536$ & 23{,}230 & 3.24 & 4.26  & 18.6 & 29.2\% & 39.7\% \\
DifIISR   & $4096\times3072$ & 15{,}944 & 3.05 & 12.00 & 26.2 & 33.4\% & 35.1\% \\
TongJi    & $8192\times6144$ & 9{,}330  & 2.85 & 20.20 & 22.1 & 21.9\% & 32.5\% \\
\bottomrule
\end{tabularx}
\end{table}

\subsection{Ground Sample Distance and Error Normalization}

Because the comparison spans four distinct image resolutions, raw pixel-based metrics are not
directly comparable: a single pixel subtends a different physical distance at each tier. Rather than
adopt Metashape's reported ground resolution --- which is back-computed from each tier's
reconstructed point distribution and therefore varies with cloud quality rather than acquisition
geometry --- we derived GSD analytically from the fixed camera geometry and sensor resolution. With
camera positions, focal length, and target standoff held constant across all tiers, the true GSD is
a function of resolution alone: 8.75~mm/px at native resolution, halving with each doubling of linear
resolution to 4.375~mm/px ($2\times$ tiers), 2.188~mm/px (DifIISR, $4\times$), and 1.094~mm/px
(TongJi, $8\times$). All physical-unit error figures below use these analytically derived values.

This normalization materially affects interpretation. In raw pixels, reprojection error appears to
degrade steeply and monotonically with enhancement aggressiveness, from 1.85~px (native) to 20.2~px
(TongJi). Once expressed in millimeters on the wall, however, the four lower-resolution tiers cluster
within a narrow 16.2--18.7~mm band, and the apparent ranking of the two most aggressive methods
inverts: TongJi (22.1~mm), operating at $8\times$ resolution where each pixel is small, is in
physical terms more accurate than DifIISR (26.2~mm), despite its far larger pixel-error figure. The
diffusion-based DifIISR is thus the least geometrically accurate method once resolution is accounted
for.

\subsection{Geometric Accuracy}

Native imagery produced the most geometrically accurate reconstruction by every measure, with the
lowest normalized reprojection error (16.2~mm) and the highest cross-view consistency (45.5\% of
points observed in three or more images). Among the enhanced tiers, the three sharing $2\times$
resolution --- UltraMax, bicubic, and SwinIR --- are directly comparable without normalization, since
they share an identical GSD. Within this group, bicubic interpolation yielded the lowest reprojection
error (17.0~mm), followed closely by SwinIR (18.6~mm) and UltraMax (18.7~mm). The two heavily
synthesized tiers, DifIISR and TongJi, separated clearly from the rest at 26.2~mm and 22.1~mm
respectively, confirming that aggressive resolution synthesis degrades geometric fidelity even after
resolution normalization.

\subsection{Tie-Point Quality and the Quantity--Quality Distinction}

Raw tie-point counts alone present a misleading picture of reconstruction value. SwinIR produced the
largest sparse cloud of any tier --- 23{,}230 points, exceeding even native resolution --- which
might suggest superior feature richness. Per-point analysis of reconstruction uncertainty, however,
reveals that much of this apparent productivity consists of poorly constrained points. Reconstruction
uncertainty is reported here as the principal quality metric precisely because it is scale-invariant:
as a dimensionless ratio of triangulation-ellipsoid axes computed against an identical camera
network, it permits direct comparison between tiers without the resolution normalization that
reprojection error requires, and the central quality finding therefore holds independently of any
assumption about pixel-to-millimeter conversion.

By this measure the tiers diverge sharply. The native cloud is effectively free of poorly
triangulated points: not a single one of its 22{,}550 tie points exceeds a reconstruction uncertainty
of 100, and only 0.9\% exceed 50. Every enhanced tier, by contrast, carries a substantial burden of
high-uncertainty points: 7.8\% for UltraMax, 21.9\% for TongJi, 29.2\% for SwinIR, 33.4\% for
DifIISR, and 41.0\% for bicubic. SwinIR's apparent tie-point advantage thus dissolves under quality
filtering --- nearly a third of its cloud comprises poorly localized points absent from the native
reconstruction, indicating that its surplus features are predominantly noise rather than recovered
detail. Bicubic exhibits the highest proportion of high-uncertainty points; lacking any mechanism to
fabricate detail, its smoothing instead produces low-gradient features that match across views yet
triangulate poorly in depth, a distinct failure mode from the fabricated texture of the generative
models.

Notably, UltraMax carries by far the lowest high-uncertainty fraction of any enhanced tier (7.8\%),
approaching native quality, while retaining cross-view consistency (42.7\% of points in $\geq$3
images) close to native's 45.5\%. Among all enhancement methods, hardware microscanning alone
produced a cloud whose quality characteristics resemble the unenhanced baseline.

\subsection{Cross-View Consistency}

The proportion of tie points observed in three or more images --- a direct measure of whether
recovered features are consistent across viewpoints rather than per-image artifacts --- followed the
same ordering as the quality metrics. Native (45.5\%) and UltraMax (42.7\%) lead; SwinIR (39.7\%) is
intermediate; and bicubic (32.0\%), TongJi (32.5\%), and DifIISR (35.1\%) trail.

These three indicators of reconstruction quality differ in their dependence on resolution.
Reprojection error is resolution-sensitive and requires GSD normalization to compare across tiers;
reconstruction uncertainty and multi-view consistency are resolution-independent, the former because
it is a dimensionless triangulation ratio and the latter because it counts observations rather than
measuring distances. The central result is that all three indicators produce the same tier ordering
--- native best, UltraMax the strongest of the enhanced tiers, and the aggressively synthesized
DifIISR and TongJi worst --- despite their differing relationships to resolution. Because the
conclusion does not rest on any single metric, and in particular because the two
resolution-independent metrics agree with the resolution-normalized one, the ranking cannot be an
artifact of the normalization procedure.

\subsection{Dense Reconstruction}

\begin{figure}[ht]
\centering
\begin{subfigure}[t]{0.32\textwidth}
    \centering
    \includegraphics[width=\textwidth]{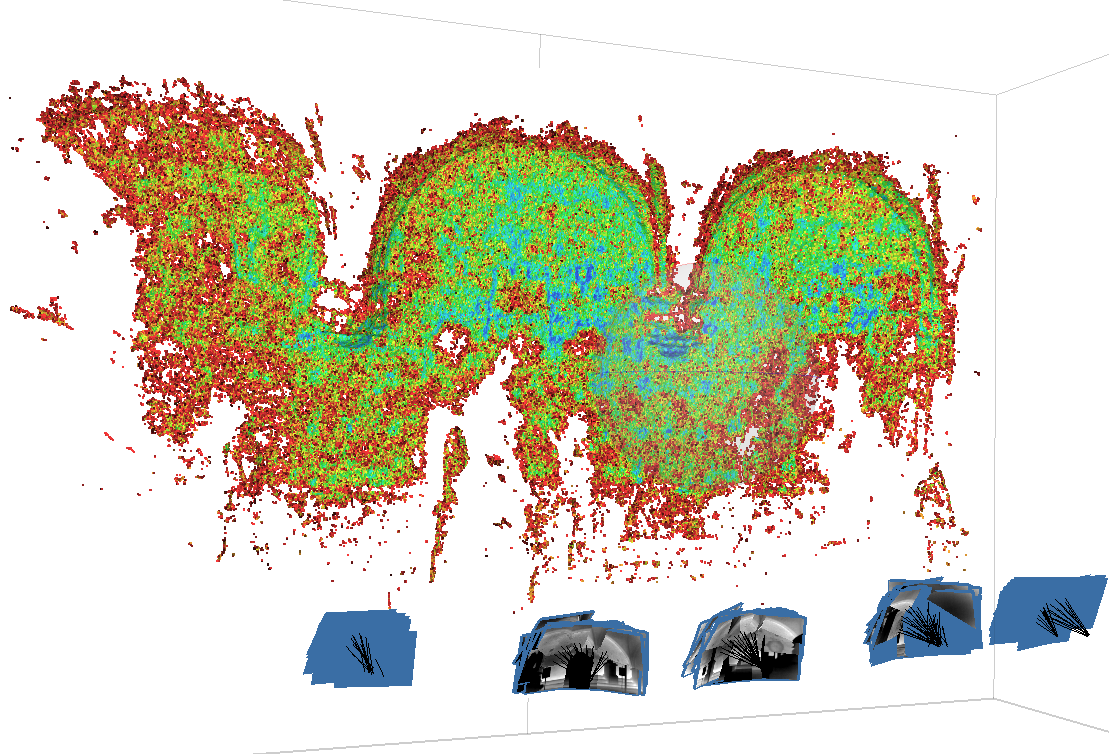}
    \caption{Native ($1024\times768$)}
    \label{fig:dense_native}
\end{subfigure}
\hfill
\begin{subfigure}[t]{0.32\textwidth}
    \centering
    \includegraphics[width=\textwidth]{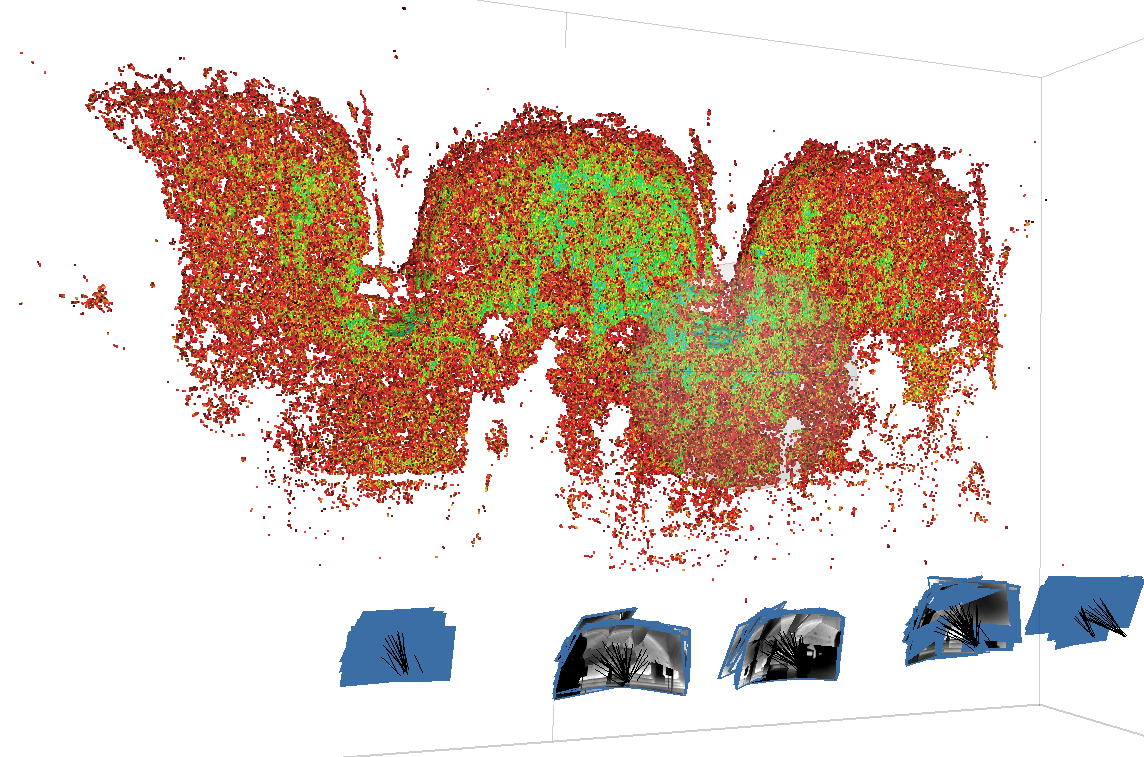}
    \caption{UltraMax ($2048\times1536$)}
    \label{fig:dense_ultramax}
\end{subfigure}
\hfill
\begin{subfigure}[t]{0.32\textwidth}
    \centering
    \includegraphics[width=\textwidth]{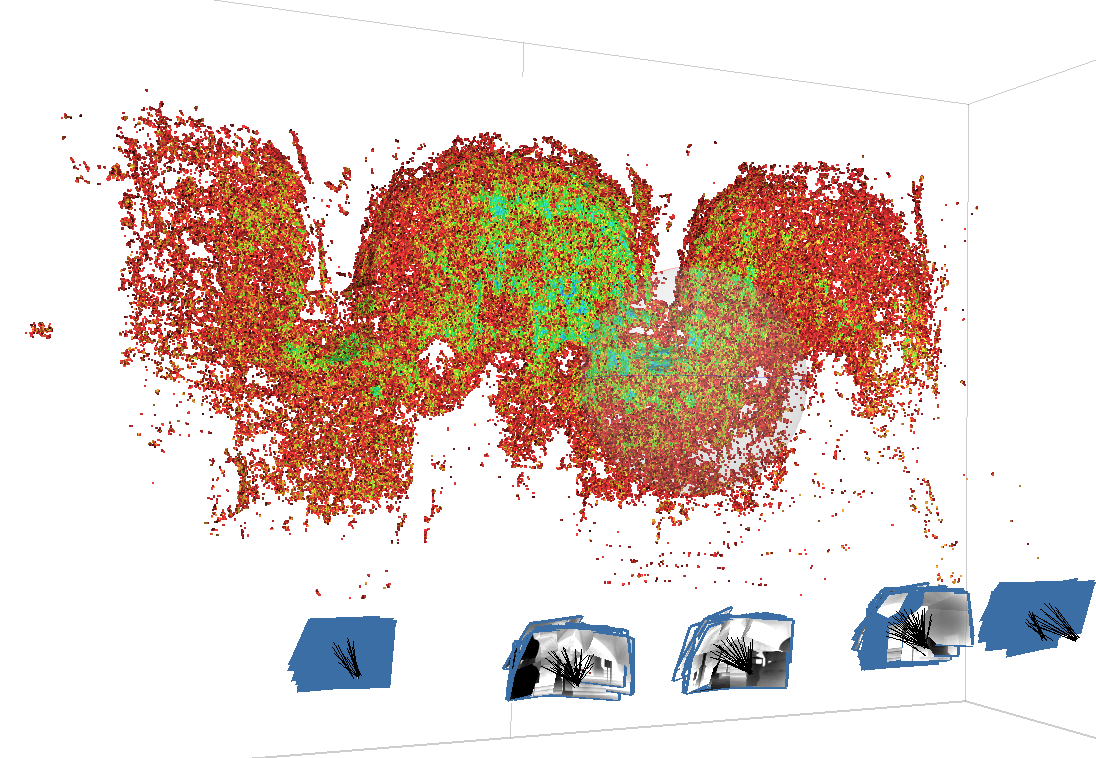}
    \caption{Bicubic ($2048\times1536$)}
    \label{fig:dense_bicubic}
\end{subfigure}

\vspace{0.75em}

\begin{subfigure}[t]{0.32\textwidth}
    \centering
    \includegraphics[width=\textwidth]{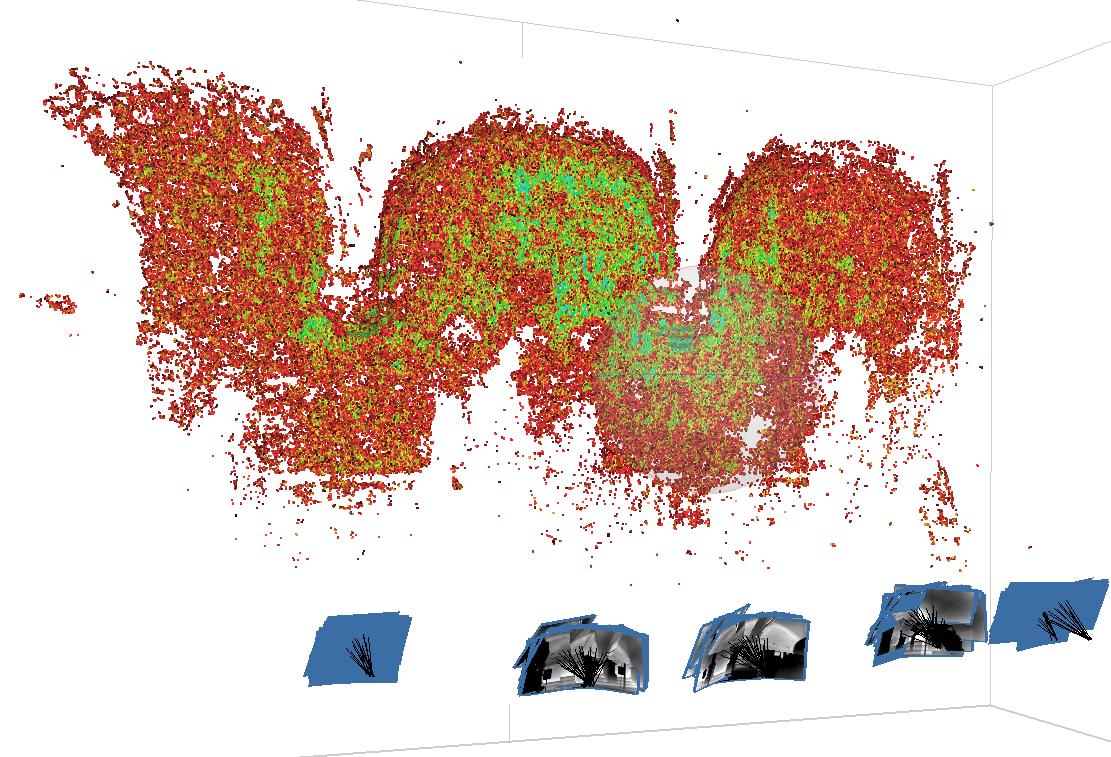}
    \caption{SwinIR ($2048\times1536$)}
    \label{fig:dense_swinir}
\end{subfigure}
\hfill
\begin{subfigure}[t]{0.32\textwidth}
    \centering
    \includegraphics[width=\textwidth]{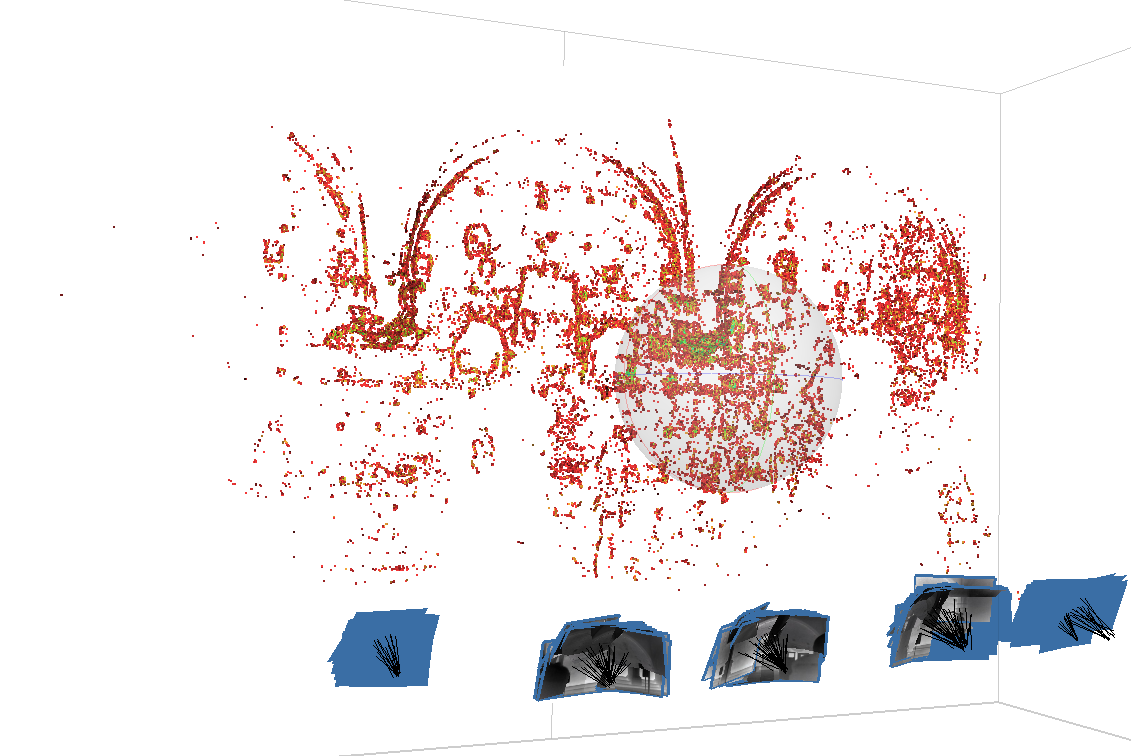}
    \caption{DifIISR ($4096\times3072$)}
    \label{fig:dense_difiisr}
\end{subfigure}
\hfill
\begin{subfigure}[t]{0.32\textwidth}
    \centering
    \includegraphics[width=\textwidth]{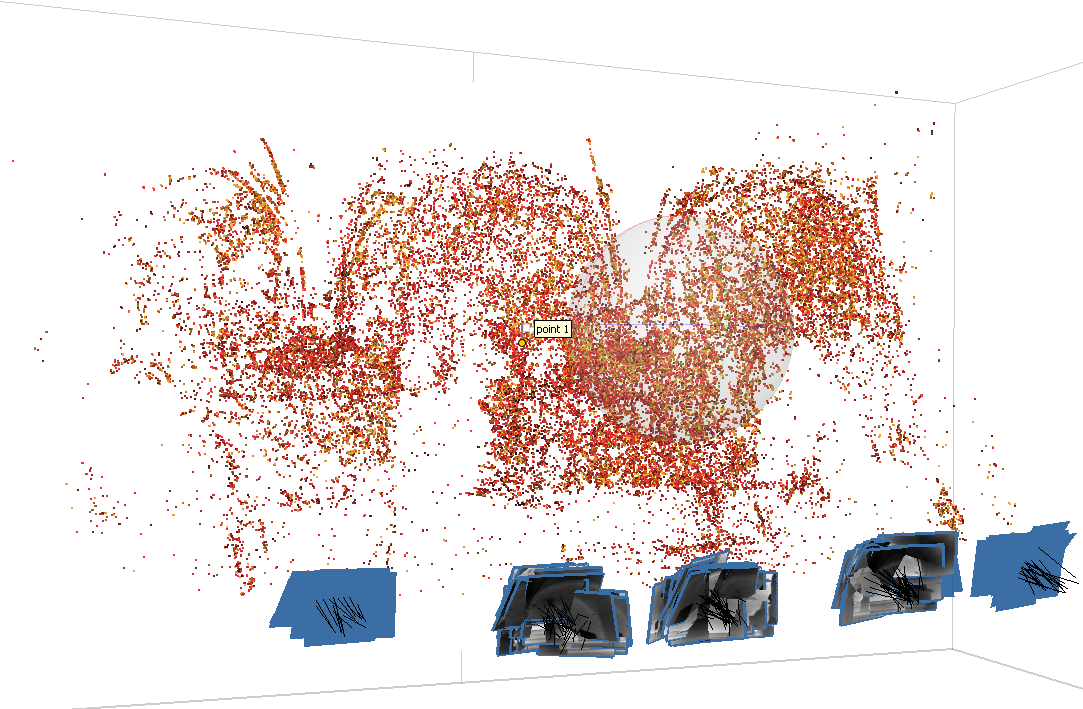}
    \caption{TongJi/DRCT ($8192\times6144$)}
    \label{fig:dense_tongji}
\end{subfigure}

\caption{The resulting dense point cloud reconstructions from each method, colored by point
confidence value (number of agreeing depth maps; red~$=1$, green~$\approx5$, blue~$\geq100$).}
\label{fig:dense}
\end{figure}

The sparse tie-point analysis characterizes the feature network that anchors each reconstruction; the
dense point cloud, built from per-pixel depth estimation across all images, reveals how each
resolution tier performs when reconstructing the continuous wall surface that heritage analysis
ultimately depends upon. Dense clouds were generated for all six tiers under identical settings with
point confidence enabled, confidence here denoting the number of depth maps in agreement at each
reconstructed point (Figure~\ref{fig:dense}). Point clouds were generated on ``ultra-high'' quality
settings, with no depth filtering. Table~\ref{tab:dense} summarizes the results.

\begin{table}[ht]
\centering
\caption{Dense reconstruction comparison. Surface completeness is the fraction of native's occupied
volume (5~cm voxels) filled by each tier; points per voxel indicates whether points are distributed
across the surface or concentrated in few locations.}
\label{tab:dense}
\small
\begin{tabularx}{\textwidth}{l R R R R R R}
\toprule
Tier & Dense pts & Mean conf. & Conf.\ $\geq$3 & Conf.\ $=$1 & Surf.\ complete & Pts/voxel \\
\midrule
Native    & 6.18\,M  & 2.43 & 32.6\% & 45.3\% & 100.0\% & 24.2 \\
UltraMax  & 19.27\,M & 1.46 & 10.1\% & 70.6\% & 93.6\%  & 80.5 \\
Bicubic   & 16.10\,M & 1.48 & 10.4\% & 70.6\% & 83.1\%  & 75.7 \\
SwinIR    & 17.67\,M & 1.42 & 9.1\%  & 72.1\% & 88.0\%  & 78.5 \\
DifIISR   & 7.16\,M  & 1.20 & 3.3\%  & 84.8\% & 18.3\%  & 152.9 \\
TongJi    & 8.03\,M  & 1.15 & 0.1\%  & 84.8\% & 26.4\%  & 118.8 \\
\bottomrule
\end{tabularx}
\end{table}

Three findings emerge, each independent of the sparse-cloud analysis yet converging on the same
conclusion.

First, dense-cloud confidence reproduces the quality hierarchy established by sparse reconstruction
uncertainty. Native achieves more than double the mean confidence of any enhanced tier (2.43 versus
1.15--1.48), and nearly a third of its points are corroborated by three or more depth maps, against
roughly one tenth for the $2\times$ tiers and almost none for the $4\times$ and $8\times$ AI methods.
Conversely, the proportion of confidence-one points --- reconstructed from a single depth map,
without corroboration --- rises monotonically with enhancement aggressiveness, from 45.3\% (native)
to 84.8\% (DifIISR and TongJi). The enhancement methods uniformly inflate point count while lowering
point confidence: UltraMax yields more than three times native's point total, yet over 70\% of those
points are uncorroborated.

Second, raw dense-point count inverts the quality ranking, as it did for sparse tie points. Ordered
by point count, native produces the fewest points of any tier and UltraMax the most --- the precise
opposite of the confidence and completeness ordering. Point count, whether sparse or dense, is
therefore not a measure of reconstruction value, a recurring theme of this study.

Third, and most striking, the surface-completeness and points-per-voxel metrics expose a qualitative
failure mode unique to the aggressively synthesized tiers. Whereas native, UltraMax, SwinIR, and
bicubic each fill 83--94\% of the reconstructed wall's spatial extent with a moderate, even point
density (24--80 points per occupied voxel), DifIISR and TongJi fill only 18.3\% and 26.4\% of that
extent despite carrying more total points than native. Their points are not distributed across the
surface but concentrated at extreme density --- 153 and 119 points per occupied voxel --- onto a
small fraction of it. Visual inspection (Figure~\ref{fig:dense}) identifies these locations as the
high-contrast structural edges: the vault ribs, arch intrados, and the sculpted frieze lines. The
smooth, low-gradient surfaces between these edges --- the plaster and ashlar of the wall itself,
where subsurface material transitions and walled-up openings would register --- are left almost
entirely unreconstructed. This is the reconstruction-domain signature of generative and high-factor
super-resolution applied to thermal imagery: the models synthesize convincing detail where strong
gradients already exist, but across the smooth interior regions they either produce no matchable
structure or introduce per-image texture too inconsistent to triangulate. For heritage documentation,
in which the continuous wall surface is the object of interest rather than its outline, this
edge-collapse renders the most aggressive AI tiers not merely lower in quality but categorically
unsuited to the task, irrespective of their high nominal point counts.

\section{Discussion}

Our photogrammetric comparison found that FLIR's UltraMax produced only a marginal improvement in
tie-point yield over the native-resolution control, though it was distinguished from the other
enhancement methods by the cleanliness of its sparse cloud: among all enhanced tiers it alone
approached native quality in reconstruction uncertainty and cross-view consistency. While this is
directionally consistent with the recovery of genuine sub-pixel detail --- as multi-frame
microscanning theory predicts --- its modest magnitude does not constitute strong evidence that
UltraMax substantially outperforms simple interpolation for heritage thermography under field
conditions. This result must be interpreted cautiously: our acquisition relied on a deliberately
loosened tripod head to approximate the operator micro-tremor that UltraMax requires, and it is
possible that this improvised jiggle did not reproduce the displacement magnitude or distribution for
which FLIR's proprietary reconstruction is optimized. Given that UltraMax has, to our knowledge,
never been independently verified in the peer-reviewed literature despite its widespread professional
adoption, the present finding should be read as a first data point rather than a definitive verdict.
Further investigation is needed across a range of controlled displacement strategies, scene types,
and thermal-contrast conditions --- ideally including precise actuator-driven sub-pixel shifts as a
positive control --- to determine whether the technology's modest benefit reflects a genuine ceiling
on hardware microscanning gains in this application, or merely the limitations of our field-improvised
acquisition. Until such work is undertaken, practitioners should regard UltraMax as a low-cost,
low-risk addition to a capture protocol rather than a substitute for optimal in-situ conditions.

The AI super-resolution methods evaluated here --- spanning a domain-agnostic transformer (SwinIR), a
domain-specific challenge winner (DRCT/TongJi-SR), and a state-of-the-art diffusion model (DifIISR)
--- yielded no measurable improvement in photogrammetric reconstruction over the native or
interpolated baselines, and by the quality metrics actively degraded it. This null result invites a
pointed question about the actual value of these tools for heritage thermography. If AI enhancement
does not produce detail consistent enough to survive geometric verification across multiple
viewpoints, and if obtaining its output requires reducing 16-bit radiometric data to 8-bit imagery
--- discarding precisely the calibrated temperature information that subsurface analysis depends upon
--- then its contribution to our use case is not merely neutral but arguably negative, trading away
measurable physical data for perceptual sharpness of no analytical benefit. The detail these models
add is optimized to satisfy a human viewer or an object-detection network, not to represent the
thermal reality of a masonry wall, and our pipeline suggests that such detail is not anchored to
consistent, multi-view physical structure. This does not mean AI super-resolution is without value in
thermography broadly; for visualization, public communication, or qualitative inspection where
radiometric precision is not required, perceptually enhanced imagery may well serve. But for
quantitative heritage documentation --- where the goal is to measure, not merely to depict --- our
findings suggest that the radiometric integrity of the native capture is worth more than any
resolution gain these models currently offer, and that effort is better directed toward optimal
acquisition conditions than toward post-hoc computational enhancement.

A necessary feature of this comparison is that it spans methods with different native output
resolutions, from the $1024 \times 768$ sensor data to the $8\times$ output of the challenge-winning
DRCT model. Comparing reprojection error across these tiers therefore relies on normalization to a
common physical unit, derived analytically from the fixed acquisition geometry. We regard this as an
inherent property of comparing super-resolution methods at their designed operating points rather
than a limitation of experimental design, and we deliberately did not resample outputs to a common
grid, as doing so would suppress the high-frequency synthesized detail whose geometric behavior is the
object of study. The robustness of our conclusions to this choice rests on the convergence noted
above: the ranking of methods is identical whether assessed by the resolution-normalized reprojection
error or by reconstruction uncertainty and multi-view consistency, both of which are independent of
resolution. A residual limitation is that the analytically derived ground sample distance assumes an
approximately constant camera-to-wall standoff across the four stations; local variation in standoff
introduces a corresponding variation in per-image GSD, though this does not affect the
resolution-independent metrics on which the principal quality conclusion depends.

\section*{AI Use Statement}
\addcontentsline{toc}{section}{AI Use Statement}

During the preparation of this manuscript, the authors used Claude.ai (Opus 4, Anthropic 2025) [34] for
the purposes of literature review, pipeline development, translation of primary sources, and
grammatical correction. The authors have reviewed and edited all output and take full responsibility
for the content of this publication.

\section*{Acknowledgements}
\addcontentsline{toc}{section}{Acknowledgements}

Special thanks to the Kinsella Innovation Fund and Carol Tohsaku for their support. Special thanks to
Professor Giorgio Verdiani of the University of Florence for his help in identifying historical
resources.


\end{document}